\documentclass[sigconf]{acmart}
\AtBeginDocument{%
 \providecommand\BibTeX{{%
 \normalfont B\kern-0.5em{\scshape i\kern-0.25em b}\kern-0.8em\TeX}}} 
\title{Augmenting Knowledge Graph Hierarchies Using Neural Transformers}

\acmConference[ECIR'24]{Goharian, N., et al. Advances in Information Retrieval. ECIR 2024. Lecture Notes in Computer Science, vol 14612. Springer, Cham.}{Springer}{Glasgow}
\setcopyright{acmcopyright}
\copyrightyear{2024}
\acmYear{2024}
\acmDOI{10.1007/978-3-031-56069-9_35}
\author{Sanat Sharma}
\email{sanatsha@adobe.com}

\affiliation{%
 \institution{Adobe Inc.}
 \country{USA}
}

\author{Mayank Poddar}
\email{mpoddar@adobe.com}

\affiliation{%
 \institution{Adobe Inc.}
 \country{USA}
}

\author{Jayant Kumar}
\email{jaykumar@adobe.com}

\affiliation{%
 \institution{Adobe Inc.}
 \country{USA}
}
\author{Kosta Blank}
\email{kblank@adobe.com}

\affiliation{%
 \institution{Adobe Inc.}
 \country{USA}
}

\author{Tracy King}
\email{tking@adobe.com}

\affiliation{%
 \institution{Adobe Inc.}
 \country{USA}
}

\begin{document}
\begin{abstract}
Knowledge graphs are useful tools to organize, recommend and sort data. Hierarchies in knowledge graphs provide significant benefit in improving understanding and compartmentalization of the data within a knowledge graph. This work leverages large language models to generate and augment hierarchies in an existing knowledge graph. For small ($<$100,000 node) domain-specific KGs, we find that a combination of few-shot prompting with one-shot generation works well, while larger KG may require cyclical generation. We present techniques for augmenting hierarchies, which led to coverage increase by 98\% for  intents and 99\% for colors in our knowledge graph.
\end{abstract}

\maketitle

\section{Introduction}

\begin{figure*}[htb]
    \centering
    \includegraphics[width=5.5in]{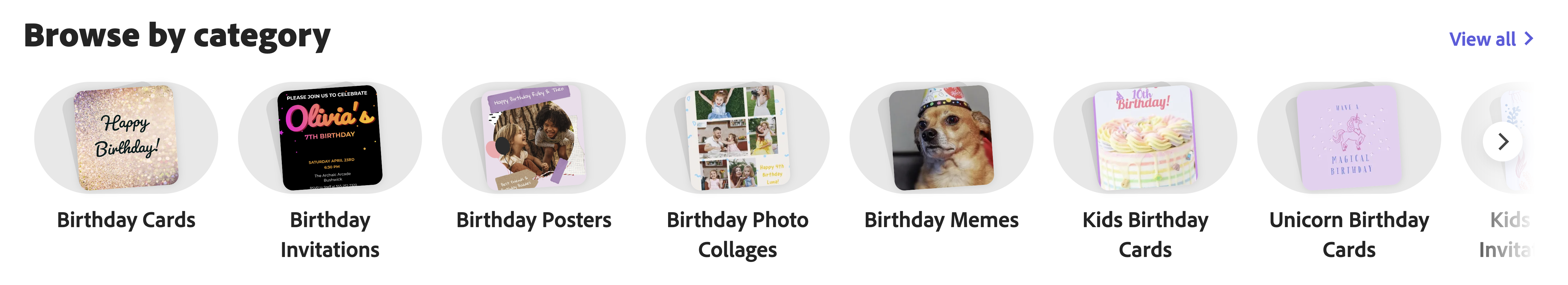}
    \caption{Adobe Express SEO page for {\em birthday} with related pages powered by the intent-based KG.}
    \label{fig:seo}
\end{figure*}

Knowledge graphs (KG) are widely used in industry to understand user behavior and provide contextual recommendations (figure \ref{fig:seo}) and search results. At Adobe, we utilize a knowledge graph  to understand  users' creative intent and  recommend Adobe assets based on the intent. While the original knowledge graph had  over 12000 intent nodes and over 100000 nodes in total, the original taxonomy  was mostly flat, lacking substantial hierarchies that could amplify the semantic significance between nodes and drive additional intent-based recommendation use cases.

In this work, we  present a novel approach to automatically generate intricate graph hierarchies in KGs by leveraging neural transformers. We enhance the structure of our graph by generating hierarchies for both intent (what an Adobe user wants to accomplish, e.g.\ create a child's birthday card or a banner for their cafe's website) and color node types, resulting in a significant increase in hierarchy coverage: 98\% for intent and 99\% for color. Hierarchies have key benefits to our users. 
	\textbf{Organizational Structure}: Hierarchical relationships  makes it easier to navigate and comprehend the KG. Hierarchies help maintain order and provide a clear understanding of how different concepts are related to each other.
  \textbf{Semantic Relationships}: Rich intent hierarchies allow us to capture the semantic relationships between concepts. They also help us unlock key  features, such as powering browse  and SEO relationships.
  \textbf{Scalability and Flexibility}: Top level categories allow for easier addition of new intents without disrupting the overall structure as the KG grows.

  \begin{figure}[htbp]
    \centering
    \includegraphics[width=4cm]{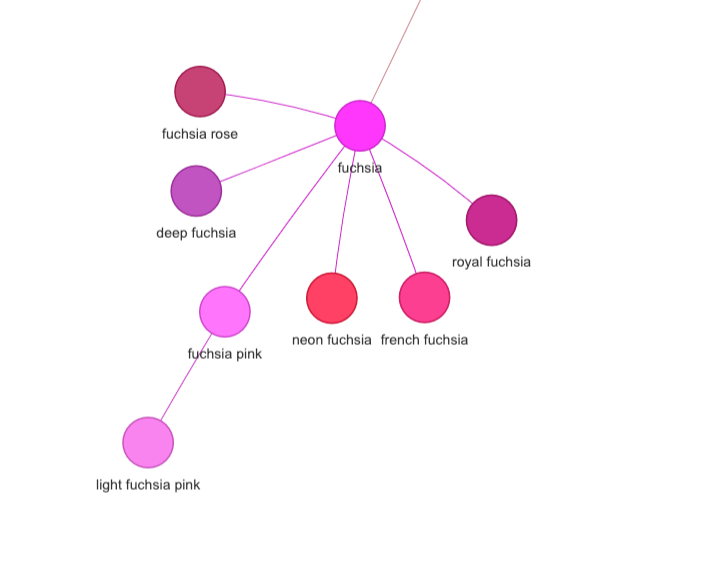} 
    \caption{Hierarchies allow us to understand and compartmentalize nodes}
    \label{fig:news}
\end{figure}


\section{Related Knowledge Graph Work}
Knowledge graphs are widely used in industry in a variety of roles, from providing social media recommendations \cite{DBLP:journals/corr/abs-1806-01973,DBLP:journals/corr/abs-1111-4503} to providing entity linking and semantic information between concepts \cite{45634,inbook}. With  recent improvements in attention-based networks, specifically large transformers \cite{openai2023gpt4,touvron2023llama}, there has been academic focus towards grounding language models with KGs \cite{pan2023unifying}, thereby providing semantic reasoning and generation based on the KG information. Recent works also investigate automated generation and completion of KGs using large transformer models \cite{meyer2023llmassisted,carta2023iterative}. They utilize language models like ChatGPT to add new nodes to subsets of the graph.
While most works focus on adding additional nodes to KGs, our work  focuses on augmenting the semantic relationships of existing nodes in the graph using large transformer models. We  generate rich hierarchies and associations inside the graph, something that is novel to the field.

\section{Approach}

First we create top level ($L_1$) categories for a specific class of nodes (e.g.\ intent or color).  $L_1$ categories can be selected by domain experts or by a language model. They need to be broad and expansive, as our aim is to transition from a general intent to a more specific one, progressing  through multiple levels.  We created the $L_1$ intent nodes by examining Adobe Express frequent queries and their intents, Adobe Stock content categories, and the Google open-source product type taxonomy \cite{googletaxonomy}. This resulted in 26 $L_1$ intent categories (e.g.\ Business and Industry, Travel, Shopping, Beauty and Wellness, Health). These overlap with standard taxonomies but comprise a subset relevant to Adobe Stock and Express users.

\begin{figure*}[htb]
    \centering
    \includegraphics[width=5in]{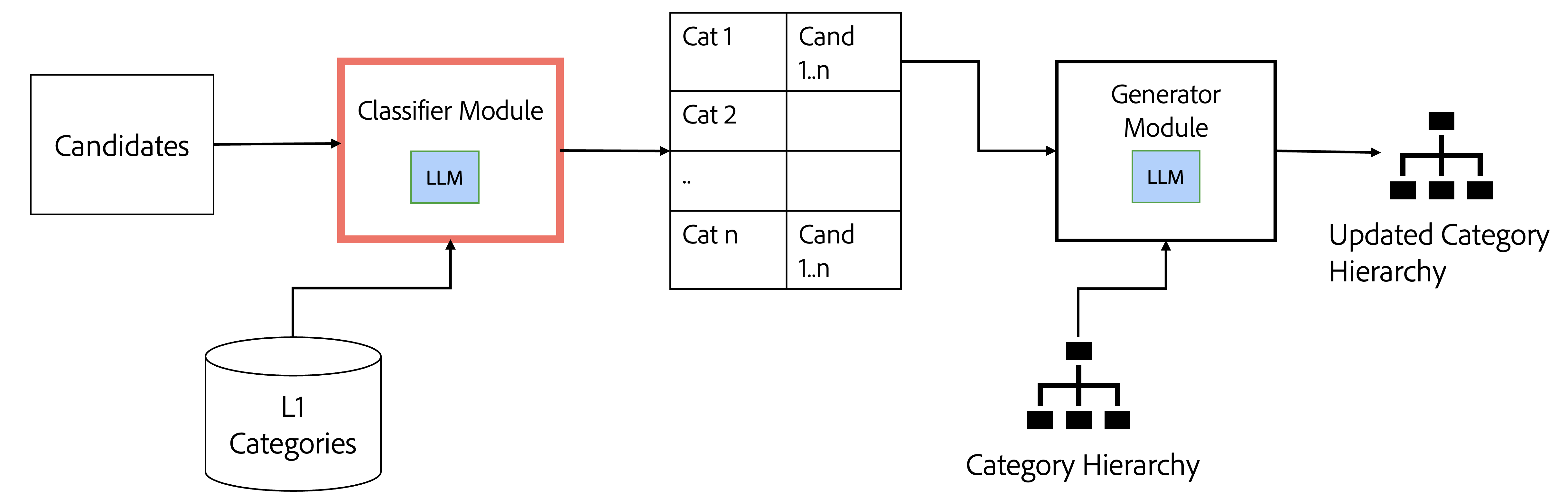}
    \caption{Hierarchy Generation Approach}
    \label{fig:full_process}
\end{figure*}

After establishing the $L_1$ categories, a classifier module assigns all KG nodes to one or more  $L_1$ categories. Then a  generator module enhances the existing hierarchy (if any) with the newly added nodes. Finally, a scalable pipeline auto-ingests the new hierarchies into the KG and queries the KG at inference time. To generate our hierarchies, we utilize two modules (Figure \ref{fig:full_process}):
\begin{enumerate}
    \item \textbf{Classifier Module}: The classifier module takes all nodes to be added to hierarchy and classifies them into one or more of the $L_1$ candidate classes.  We found large language models to be better at few-shot classification than their smaller variants.
    \item \textbf{Generator Module}: The generator module  runs in a loop for each $L_1$ category. The generator takes the existing hierarchy for that $L_1$ category (just  the $L_1$ node  if no hierarchy exists) and adds all the candidate nodes to generate the updated category hierarchy. We found one-shot hierarchy generation using large language models to be the best approach  (\S \ref{sec:gen}).
\end{enumerate}

 \subsection{Few-Shot Prompting}
 \label{sec:prompt}

In order to use the classification module, we do few-shot learning in which we provide the language model (GPT4 with a 32K context \cite{openai2023gpt4}) with a few classification examples and a strict prompt. A sample prompt we used is {\em “You are a taxonomist, classify the given node to one or more of the provided categories. If you think the category should be its own thing, return Other. Please return a dictionary every time.”}\/
With the prompt, we  provide  a few sample nodes, the categories and an output prediction. Based on a few rounds of samples, we then provide the true candidates for classification to the model. Similar to other approaches in the industry \cite{parnami2022learning,song2022comprehensive}, we see a significant boost of 12\% in accuracy by doing few-shot learning compared to zero-shot classification.


\subsection{Generation Module}
\label{sec:gen}

Once the candidates are categorized, we experimented with two techniques in the generation module to create the updated hierarchy. \textbf{Cyclical Generation}: Generate each level of the category in a loop. This means that $L_2$ level nodes are added first, then $L_3$ and so on. This is needed when the existing hierarchy is large and cannot fit in the model context.
 \textbf{One-Shot Generation}: All candidate nodes are added to the hierarchy in a single pass, without any cyclical generation. We found that this approach  produced better results with smaller ($<$100,000 node) taxonomies.

\textbf{Cyclical Generation}
In the cyclical generation approach, we follow a cyclical pattern to generate each level of the hierarchy for an $L_1$ category and its children.

\begin{enumerate}
    \item From the candidate set of nodes at level $L_i$, classify nodes that belong to that level. Utilize the existing nodes at that level (i.e.\ any nodes already present in the hierarchy) to help the language model via a few-shot approach. 
    \item The nodes not categorized to be part of level $L_i$ will  become part of lower levels ($L_{i+1..n}$). 
    \item Pass the remaining nodes as well as the existing hierarchy to the generator module to create the new updated hierarchy for level $L_i$. The generator module will attempt to categorize and place each of the remaining nodes under one of the $L_i$ nodes. 
    \item For each of the $L_i$ nodes and their hierarchy, repeat the process in a recursive manner to fine-tune the hierarchies. The process stops when either a specified depth ($L_i$) is reached or all nodes have been added to the hierarchy.

\end{enumerate}

\begin{figure*}[htb]
    \centering
    \includegraphics[width=5in]{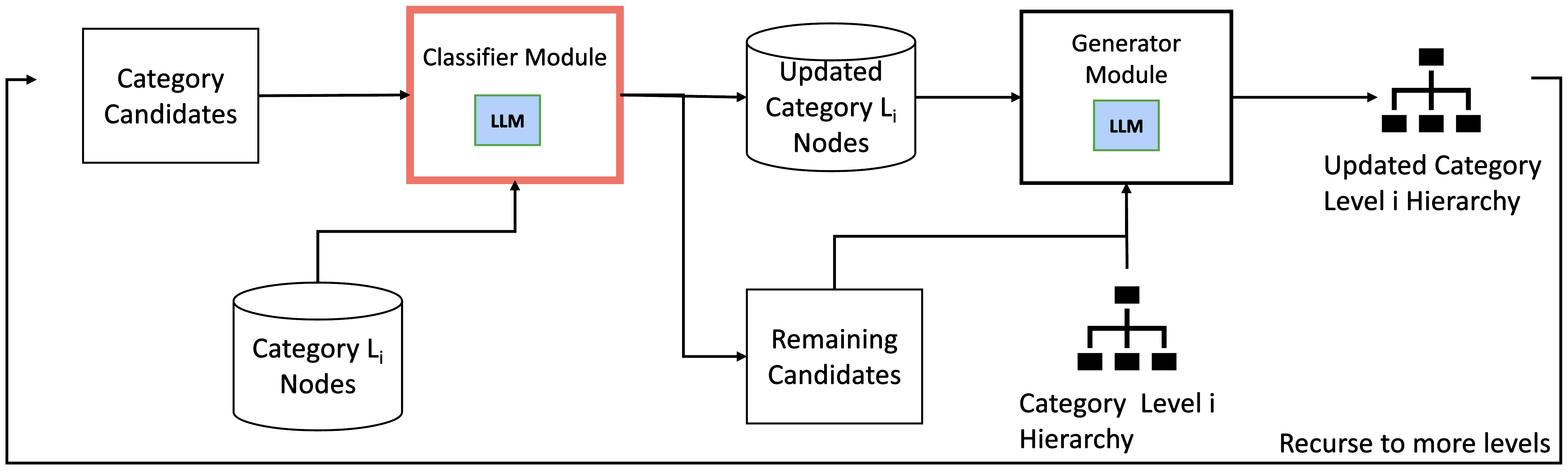}
    \caption{Cyclical Generation requires classification of nodes and addition at each level}
    \label{fig:cyclical}
\end{figure*}

While the cyclical approach is useful, especially for larger graphs, we saw several drawbacks with it when generating our intent hierarchies.
\begin{enumerate}
 \item\textbf{Error Propagation}: LLMs can  hallucinate or generate incorrect structured content. Having multiple steps in the generation process can lead to error propagation through the chain. This is the biggest issue with a recursive approach. 
 \item\textbf{The Other Conundrum}: LLMs are bad at placing nodes into the “Other” category. This means that most nodes were assigned into a level’s category (L$_{i}$) rather than being placed into the Other category (to be a part of the L3 and lower levels). 
 \item\textbf{Order Importance}: Whether we pass nodes in a batch or one at a time for classification, the order of nodes  plays a huge difference in the categorizing. For example, if “birthday party” was categorized as an L$_{i}$ first and then the node “birthday” is shown to the LLM, it often incorrectly categorizes “birthday” as a child of “birthday party” due to their similarities. Additional checks and another overall pass is required to fix the categorizations. One-shot generation (see below) alleviates these issues.
 \end{enumerate}

\textbf{One-Shot Generation}
In the one-shot generation process, we provide all the nodes to be added as well as the full existing hierarchy to the LLM and allow it to generate the $L_2$, $L_3$ and lower categories with just a few examples provided. This approach  worked well when there was an existing, partial hierarchy  in place to  guide the language model’s intuition. If a large number of candidates need to be added to the hierarchies, batched generations followed by an overall pass where the language model has the chance to correct any errors is utilized. For our domain-specific use cases, we found full, one-shot generation to be more viable since our taxonomy is relatively small ($<15000$ intent nodes and $<3000$ color nodes).

\begin{figure}[htbp]
    \centering
    \includegraphics[width=5cm,height=5cm]{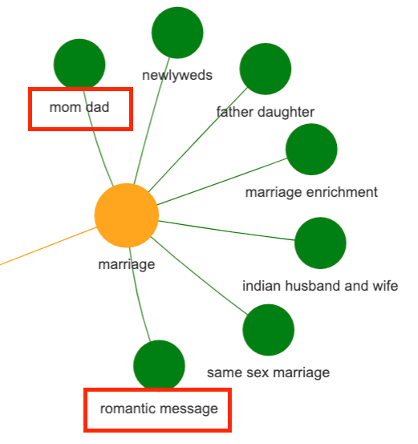} 
    \caption{LLMs can sometimes confuse similar semantic meaning with parent-child relationship}
    \label{fig:fail}
\end{figure}

\begin{figure*}[htb]
    \centering
    \includegraphics[width=5in]{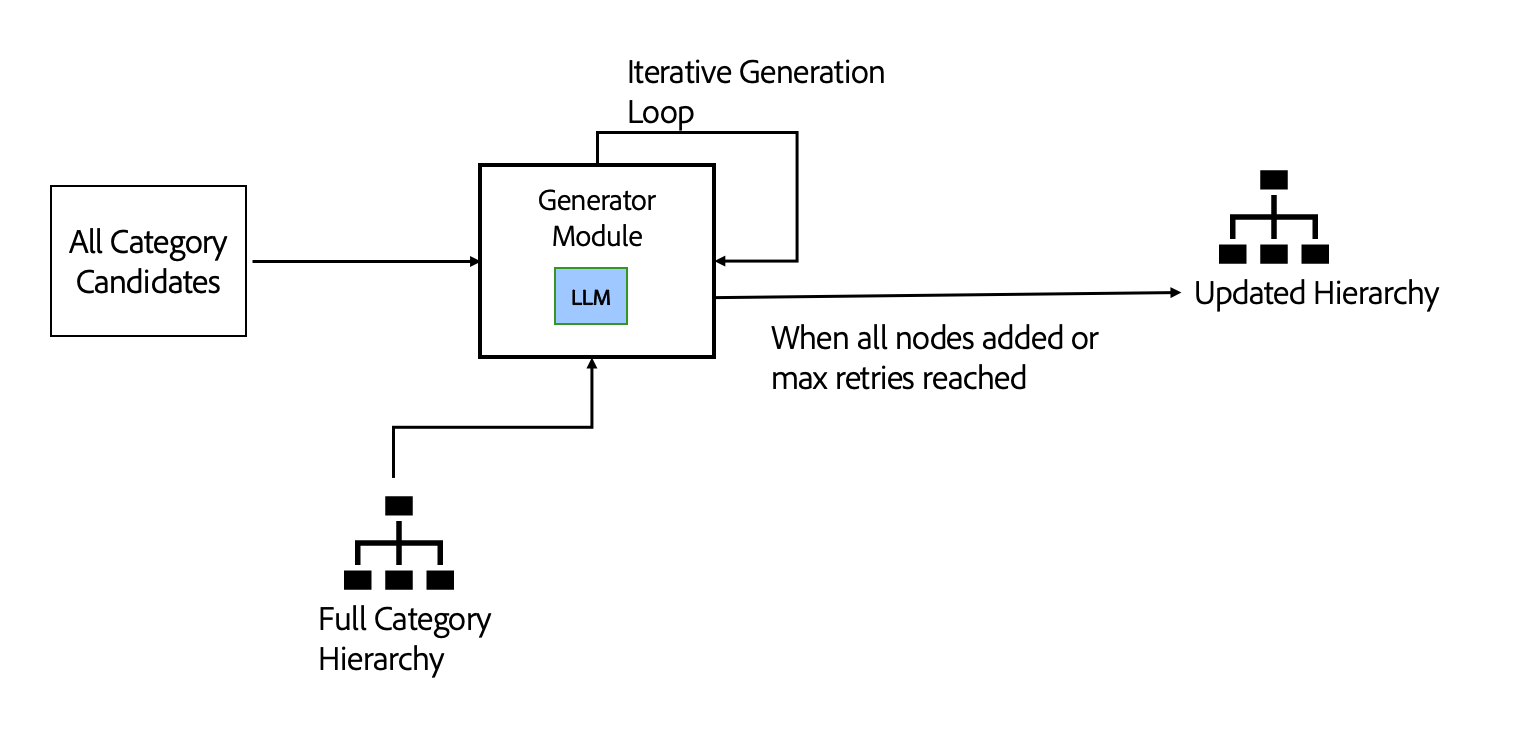}
    \caption{One Shot Generation is an iterative generation process where graph hierarchies are added in one go by prompting the LLM}
    \label{fig:one_shot}
\end{figure*}

\begin{figure*}[htb]
    \centering
    \includegraphics[width=4in]{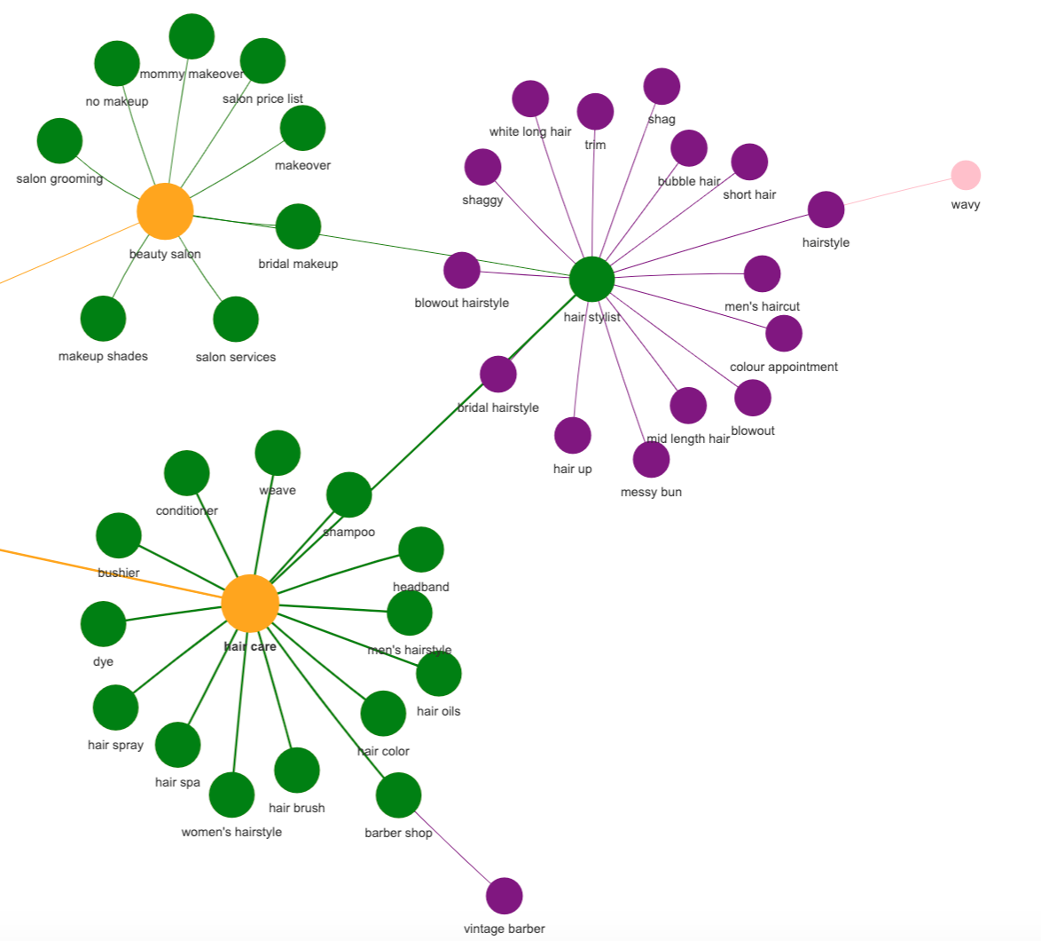}
    \caption{Subset of L4-L6 hierarchies for Beauty and Wellness category. Each node can have multiple parents.}
    \label{fig:one_shot}
\end{figure*}

\begin{figure*}[htb]
    \centering
    \includegraphics[width=5in]{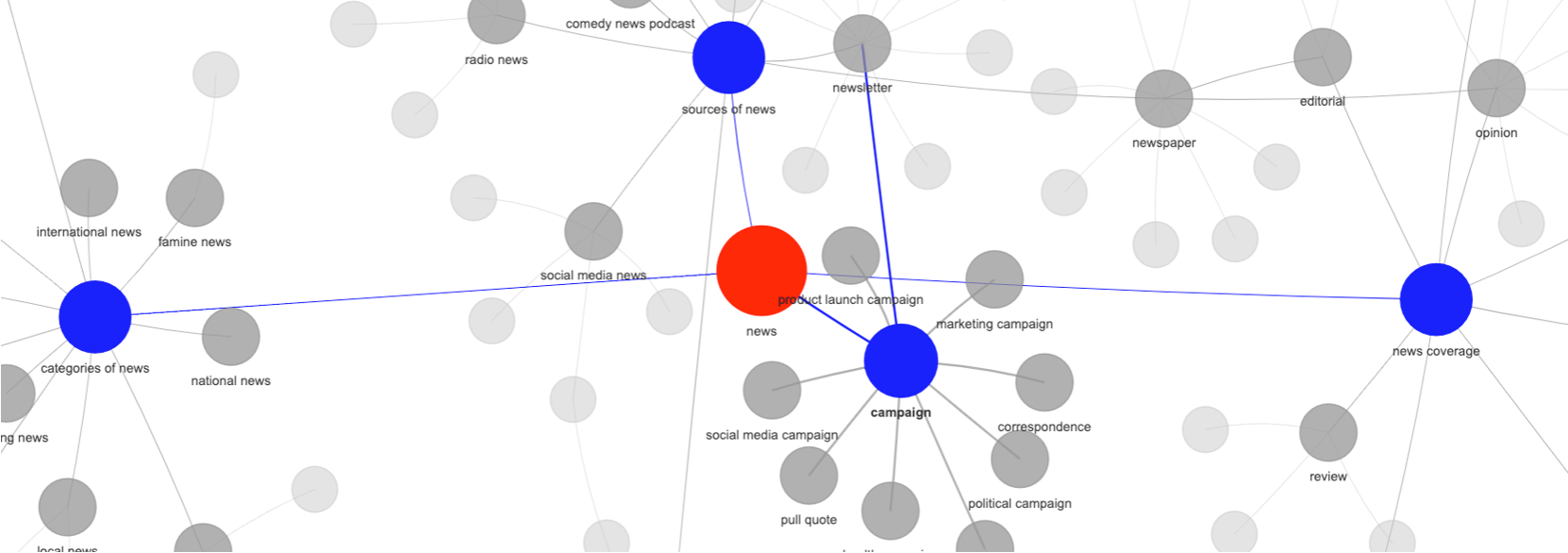}
    \caption{Subset of L1-L2 hierarchies for News category.}
    \label{fig:news}
\end{figure*}

\textbf{Updating the Graph} For new intents, the above approach is used to integrate them into the KG. When creating intents for a new domain (e.g.\ Adobe app tools), a subgraph is generated for the new domain and then merged into the existing KG. One  algorithm improvement, suggested by an anonymous reviewer, is to examine each subgraph in the generated graph and query the LLM as a third step if it looks good. This uses the LLM for evaluation and updating, instead of generating.

\textbf{Key Failure Cases}

One of the key failure cases we saw with using LLM for hierarchical generation is confusion of similar semantic nodes. As shown in Fig. \ref{fig:fail}, LLMs can sometimes get confused with related concepts and ascribe a parent-child relationship. While \textit{mom dad} and \textit{romantic message} are related to the concept of love, they should be children of \textit{marriage}. We found few-shot prompting and  review pass-throughs (asking LLM to find flaws in the hierarchy once generated) to be the most effective techniques in preventing and correcting these errors.

\section{Hierarchical KG Evaluation and Conclusions}

Statistics on the hierarchical KG generated using one-shot generation and few-shot prompting are summarized below (`Lower' indicates nodes in $L_5$ or below categories). 

\begin{table*}[t]
    \centering
    \caption{Augmentation of Graph hierarchies}
    \begin{tabular}{|l|lll|lllll|}
    \hline
         KG & Nodes & In Hierarchy Before & In Hierarchy Now & $L_1$ & $L_2$ & $L_3$ & $L_4$ & $l_5$ and Lower\\\hline
         Intents&12385&956&12339&25&904&4684&4961&3195\\
         Colors&328&12&328&12&772&237&7&0\\
         \hline
    \end{tabular}
\end{table*}

We evaluated the KG hierarchies through a human-in-the-loop approach. We provided  a graphical interface to identify nodes that are incorrectly positioned and to offer suggestions for enhancement. We engaged  16 Adobe-internal domain experts to  review the hierarchies within each $L_1$ category for both intent and color nodes. The hierarchies were found to be relevant  $>$95\% of the time. Lower levels were spot-checked for accuracy. Identified errors were then manually corrected. 

The ultimate evaluation will be in leveraging the KG hierarchies in search and recommendation features. The non-hierarchical graph  already provides related search style links between Express SEO pages (figure \ref{fig:seo}) and  powers null and low recovery in Adobe Express by mapping queries and  templates to intents. These use cases will be enhanced by using the hierarchy to provide additional links, to type the links, and to provide back-off through the hierarchy. The Express SEO color pages represent the first user-facing application of the hierarchical graph.

\bibliographystyle{splncs04}
\bibliography{ECIR_prefs}

\end{document}